\documentclass[runningheads]{llncs}

\usepackage[mobile]{eccv}
\usepackage{url}            % simple URL typesetting
\usepackage{booktabs}       % professional-quality tables
\usepackage{amsfonts}       % blackboard math symbols
\usepackage{amsmath}        % 
\usepackage{nicefrac}       % compact symbols for 1/2, etc.
\usepackage{microtype}      % microtypography
\usepackage{xcolor}         % colors
\usepackage{gensymb}
\usepackage{amssymb}
\usepackage{subcaption}
\usepackage{bm}
\usepackage{bbm}
\usepackage{graphicx}
\usepackage{multirow}
\usepackage{subcaption}

\usepackage{eccvabbrv}

\usepackage{graphicx}
\usepackage{booktabs}
\usepackage{multirow}
\usepackage[table]{xcolor}

\usepackage[accsupp]{axessibility}  % Improves PDF readability for those with disabilities.

\usepackage{hyperref}

\usepackage{orcidlink}

\begin{document}

\title{InpaintSLat: Inpainting Structured 3D Latents \\ via Initial Noise Optimization} 

% \titlerunning{InpaintSLat}

\author{Jaeyoung Chung \and Suyoung Lee \and Kyoung Mu Lee}

\authorrunning{J. Chung et al.}

\institute{Seoul National University, Seoul, South Korea. \\
\email{\{robot0321, esw0116, kyoungmu\}@snu.ac.kr}
}

\maketitle

\begin{figure*}[h]
    \vspace{-6mm}
    \includegraphics[width=\textwidth]{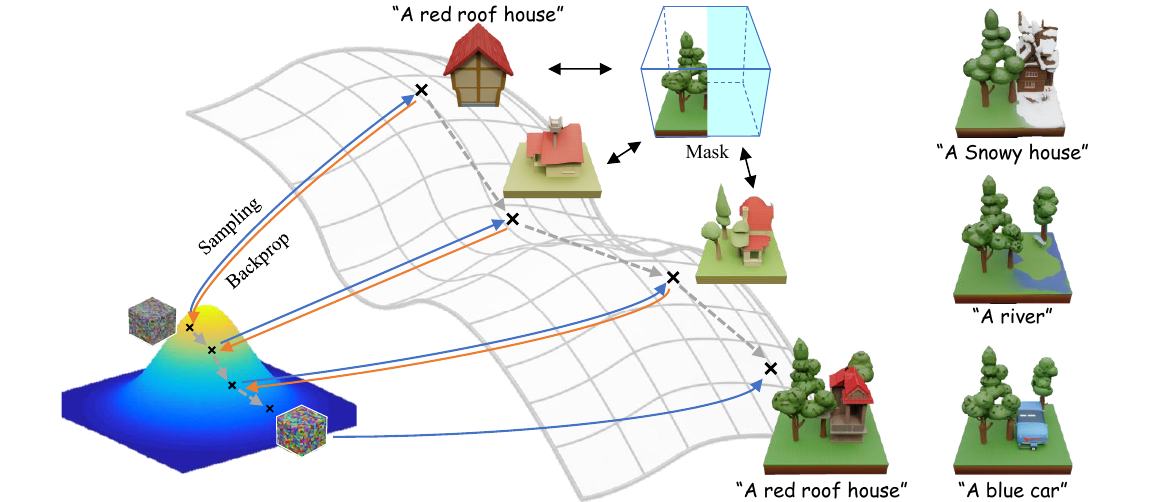}
    \caption{
        \textbf{Teaser.} 3D asset inpainting via initial noise optimization.
    }
    \label{fig:teaser}
    \vspace{-10mm}
\end{figure*}
\begin{abstract}
We present a training-free approach for controllable 3D inpainting based on initial noise optimization. In the structured 3D latent diffusion framework, we observe that the underlying geometric structure is established during the early stages of the diffusion process and exhibits high sensitivity to the initial noise. 
Such characteristics compromise stability in tasks like inpainting and editing, where the model must ensure strict alignment with the existing context while synthesizing a new structure.
In this paper, we introduce a strategy to optimize the initial noise within the structured 3D latent diffusion framework, ensuring high-fidelity 3D inpainting.
Specifically, we update the initial noise by leveraging a backpropagation approximation grounded in the rectified flow model, with the spectral parameterization specially designed for robust and efficient structured 3D latent optimization.
Experiments demonstrate consistent improvements in contextual consistency and prompt alignment over representative training-free inpainting baselines, establishing initial noise control as an independent dimension for 3D inpainting, orthogonal to conventional sampling trajectory manipulation.
\end{abstract}
\section{Introduction} \label{sec:intro}
%% 여는글: Why 3D inpainting?
Recent breakthroughs in 3D generative modeling, most notably structured latent frameworks like TRELLIS~\cite{xiang2025structured}, have significantly advanced our ability to synthesize high-fidelity 3D assets from various modalities. However, extending these capabilities to 3D inpainting—the task of completing missing regions of a 3D object while preserving existing structures—remains a formidable challenge. Unlike 2D image inpainting, where the model primarily predicts pixel intensities across a dense, fixed grid, 3D inpainting must simultaneously resolve both geometry and appearance. This introduces a fundamental difficulty: while 2D pixels occupy the entire image space by definition, 3D generative models must determine the occupancy of the missing regions, essentially deciding not only what color to fill but also where to place the surface itself. This additional structural requirement demands a much tighter coupling between the generative prior and the contextual constraints.

Existing training-free approaches for controllable generation largely rely on steering the sampling trajectory, a strategy successfully pioneered in 2D diffusion models through methods like RePaint~\cite{lugmayr2022repaint} and SDEdit~\cite{meng2021sdedit}. When directly applied to 3D structured latents, however, these trajectory-based manipulations often exhibit significant limitations. In 3D generation, the global geometric layout is established during the early stages of denoising and is heavily dictated by the initial noise seed. When the random initial seed is poorly aligned with the target structure, forcing the sampling trajectory toward the constraints often results in a lack of robustness. This often leads to structural collapses, where the model struggles to synthesize textures onto fragmented or incoherent geometric foundations, resulting in artifacts that the pretrained prior cannot naturally resolve.

\begin{figure*}[t]
    \includegraphics[width=1.0\textwidth, height=0.5\textwidth]{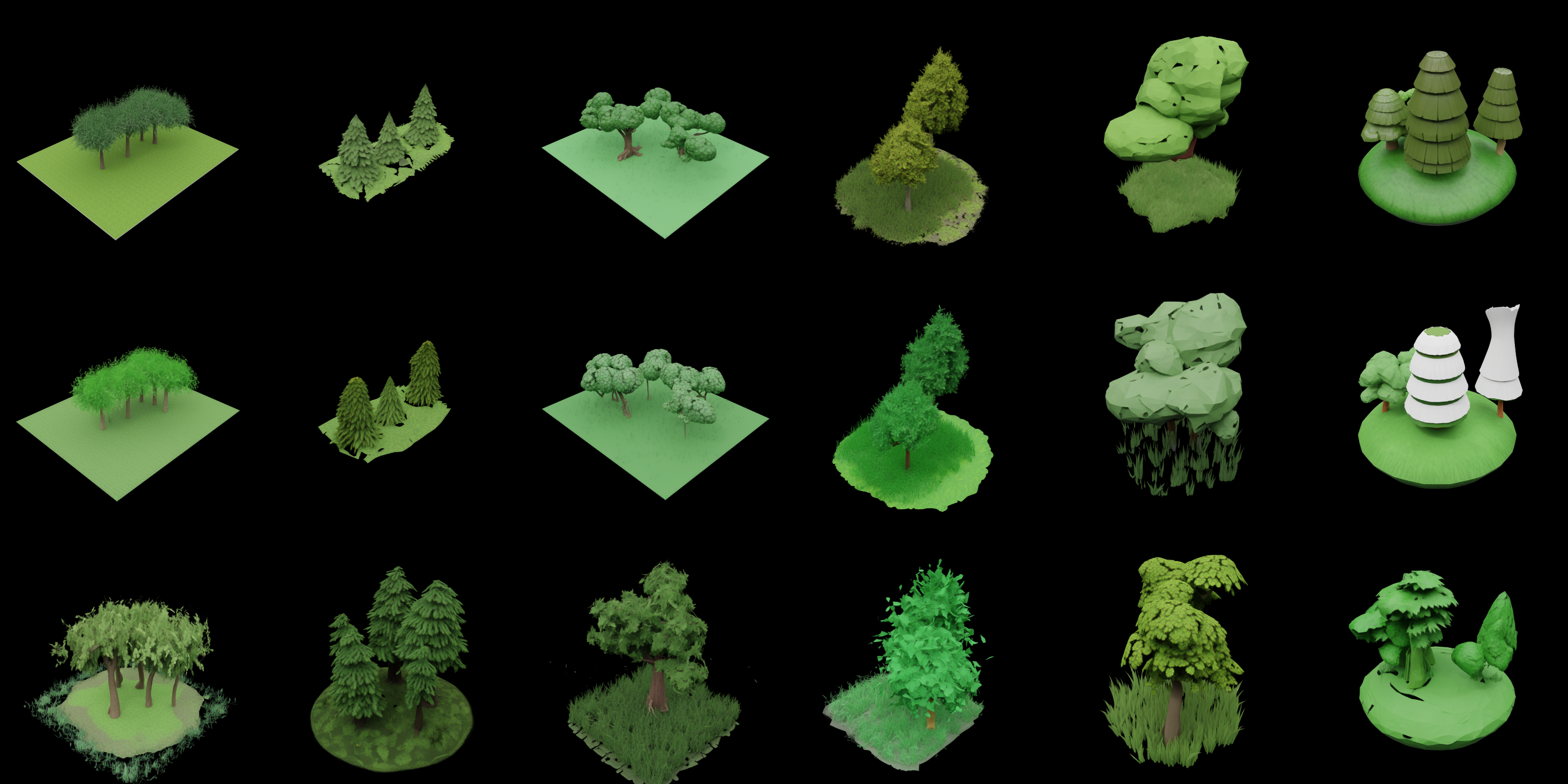}
    \caption{
        The output of TRELLIS is highly sensitive to initial noise, showing various types of 3D.
    }
    \label{fig:sensitive_initialseed}
    \vspace{-3mm}
\end{figure*}

To understand the root of this sensitivity, we investigate the relationship between the initial noise and the final synthesized 3D structure. As illustrated in \cref{fig:sensitive_initialseed}, we observe that the mapping from the initial latent space to the generated output exhibits two distinct behaviors: (1) small perturbations in the initial noise result in locally smooth variations in the output, but (2) large deviations in the seed lead to drastic, non-linear shifts in the fundamental geometric backbone. These observations suggest that the quality and reliability of 3D inpainting are inherently bound to the suitability of the starting noise. Consequently, instead of merely correcting the path, we explore a more principled direction: identifying an optimal initial noise seed that is inherently compatible with both the preserved context and the target prompt.

In this work, we propose a suite of technical innovations to enable stable and efficient initial seed optimization for 3D inpainting. To manage the computational complexity of backpropagating through 3D diffusion steps, we introduce an approximate backpropagation strategy based on rectified flow modeling. To better capture the multi-scale nature of 3D shapes, we employ spectral parameterization (frequency-based updates) for the geometric components of the structured latent, which ensures smoother optimization of the global structure. Furthermore, we incorporate a distributional regularization loss to maintain the optimized noise within the i.i.d. Gaussian manifold of the pretrained prior, ensuring generative robustness. Crucially, our seed optimization framework is orthogonal to, and thus compatible with, existing trajectory-steering methods. By refining the initialization, we provide a coherent structural anchor that can either stand alone or be paired with per-step guidance to achieve enhanced inpainting quality.

Our contributions are summarized as follows:
%%%
\begin{itemize}
\item We propose a training-free 3D inpainting framework that optimizes the initial structured latent seed to ensure geometric consistency and structural integrity.
\item We introduce a technical suite for stable 3D latent refinement, including approximate backpropagation via rectified flow, spectral parameterization for 3D geometry, and Gaussian regularization.
\item We demonstrate that seed optimization is orthogonal to existing trajectory-based guidance, offering a versatile and robust approach that significantly reduces geometric failures compared to traditional sampling manipulation.
\end{itemize}

\section{Related Work} \label{sec:relatedwork}

% \subsection{3D Generative Models and Inpainting}
% 1. about 3D generative models
% 2. bare inpainting-model -> training inpainting-dedicated model (X)
% 3. training-free approach

% \subsection{Training-Free Approaches for 2D Inpainting}
% 1. sampling guidance (including RePaint?)
% 2. Blending (RePaint, MultiDiffusion)
% 3. initial-seed approaches

\subsection{3D Generative Models and Inpainting}
\paragraph{3D Generative Models.}
Text-to-3D generation has progressed rapidly with the introduction of score distillation methods such as DreamFusion~\cite{poole2022dreamfusion}, Magic3D~\cite{lin2023magic3d}, and ProlificDreamer~\cite{wang2023prolificdreamer}, which leverage pretrained 2D diffusion priors to optimize neural radiance fields or meshes. More recently, native 3D diffusion models have emerged that directly model structured 3D representations. XCube~\cite{ren2024xcube} and TRELLIS~\cite{xiang2025structured} integrate diffusion with voxelized or structured latent representations, enabling scalable 3D generation with explicit geometric modeling. These approaches establish structured 3D latent diffusion as a practical foundation for controllable 3D synthesis.
\paragraph{3D Completion and Editing.}
3D completion has traditionally been studied as shape completion from partial observations, including methods such as PCN~\cite{yuan2018pcn} and subsequent point cloud completion models. Neural radiance field editing approaches further enable localized geometry or appearance modification via per-scene optimization~\cite{yuan2022nerf}. However, these methods typically require supervised training or scene-specific optimization and are not designed for training-free inpainting under a pretrained 3D diffusion prior. To date, there is no established training-free inpainting framework built upon structured 3D latent diffusion models such as TRELLIS or XCube.

\subsection{Training-Free Approaches for 2D Inpainting}
\paragraph{Sampling-Based Guidance.}
Training-free diffusion control in 2D is commonly achieved by modifying the sampling trajectory. RePaint~\cite{lugmayr2022repaint} introduces masked resampling to enforce region consistency during diffusion. SDEdit~\cite{meng2021sdedit} enables editing by adding and removing noise while preserving structure. Diffusion Posterior Sampling (DPS)~\cite{chung2022diffusion} and ILVR~\cite{choi2021ilvr} incorporate measurement consistency or low-frequency constraints to guide generation during sampling. These methods operate by injecting constraints into intermediate denoising steps.
\paragraph{Spatial Blending and Aggregation.}
MultiDiffusion~\cite{bar2023multidiffusion} performs patch-wise aggregation to enforce regional consistency across spatial domains, while Blended Diffusion~\cite{avrahami2022blended} combines generated and original content through spatial blending. Such approaches control generation by manipulating intermediate denoised predictions rather than modifying model parameters.
\paragraph{Initial Noise and Latent Optimization.}
Recent works demonstrate that diffusion outcomes are highly sensitive to the initial noise. Null-Text Inversion~\cite{mokady2023null} optimizes latent variables for real-image editing without retraining the model. SONIC~\cite{baek2025sonic} explicitly optimizes the initial latent to improve conditional alignment in diffusion models. These results suggest that initialization provides an alternative axis for diffusion control beyond trajectory manipulation.

Despite these advances in 2D diffusion, initialization-based control has not been systematically explored in structured 3D latent diffusion models. Our work bridges this gap by introducing seed optimization for training-free 3D inpainting under geometric constraints.
\section{Method} \label{sec:method}

% \subsection{Structured Latent Model}
% 1. TRELLIS (preliminary)의 sparse structure / slat 생성 방식 소개 (notation을 TRELLIS와 맞춰서)

% \subsection{Optimize Initial Noisy Latent by Approximated Gradient}
% 1. Gradient approximation in 3D Rectified Flow model (inspired by SONIC)

% \subsection{Optimization Strategies for SLAT-conditioned recon}
% 1. Frequency optimization for sparse structure / spatial basis for SLAT

% \subsection{Optimization Strategies for diversity and quality for Inpainting}
% 1. 여러번 돌아야 되서 시간이 많이 걸림 -> early step optimization 방식 도입
% 2. distribution matching loss (Gasussian dist.가 되도록 mean, std, skew, kurtosis 맞추기)

\subsection{Structured Latent Generation in TRELLIS}
\label{sec:trellis_prelim}

TRELLIS~\cite{xiang2025structured} represents a 3D asset $\mathcal{O}$ using a structured latent
\begin{equation}
\mathbf{z} = \{(z_i, p_i)\}_{i=1}^{L}, \quad 
z_i \in \mathbb{R}^C, \quad 
p_i \in \{0,\dots,N-1\}^3,
\end{equation}
where $p_i$ denotes active voxel coordinates in an $N^3$ grid and $z_i$ is the local latent attached to voxel $p_i$.
The active voxel set defines coarse geometry in sparse structure, and $\{z_i\}$ encode fine geometric and appearance details.
The generation process is consist in two process: sparse structure generation and structured latent generation.
A rectified flow model $\mathcal{G}_S$ denoise in feature grid $S \in \mathbb{R}^{D \times D \times D \times C_S}$, and it is decoded into the discrete grid $O$, and further converted back to active voxels ${p_i}^L_{i=1}$ as the final sparse structure.
Conditioned on $\{p_i\}$, a second rectified flow model $\mathcal{G}_L$ generates the local features $\{z_i\}$ directly on sparse voxel coordinates, building structured latent $\mathbf{z}$.
Finally, the structured latent $\mathbf{z}$ is decoded into different 3D representations (e.g., Gaussian splatting, radiance fields, meshes) via task-specific decoders $\mathcal{D}$.
In this model, both $\mathcal{G}_S$ and $\mathcal{G}_L$ are trained using Conditional Flow Matching (CFM):
\begin{equation}
x(t) = (1-t)x_0 + t\epsilon, \quad \mathcal{L}_{\text{CFM}} = \mathbb{E}_{t,x_0,\epsilon} \|v_\theta(x,t) - (\epsilon - x_0)\|_2^2,
\label{eq:RectifiedFlowModel}
\end{equation}
where $v_\theta$ is time-dependent vector field, and $\epsilon$ is estimated noise.
The linear interpolation structure of rectified flow yields locally linear denoising dynamics, which later enables efficient approximation of gradients with respect to the initial latent.

\begin{figure*}[t]
    \includegraphics[width=1.0\textwidth]{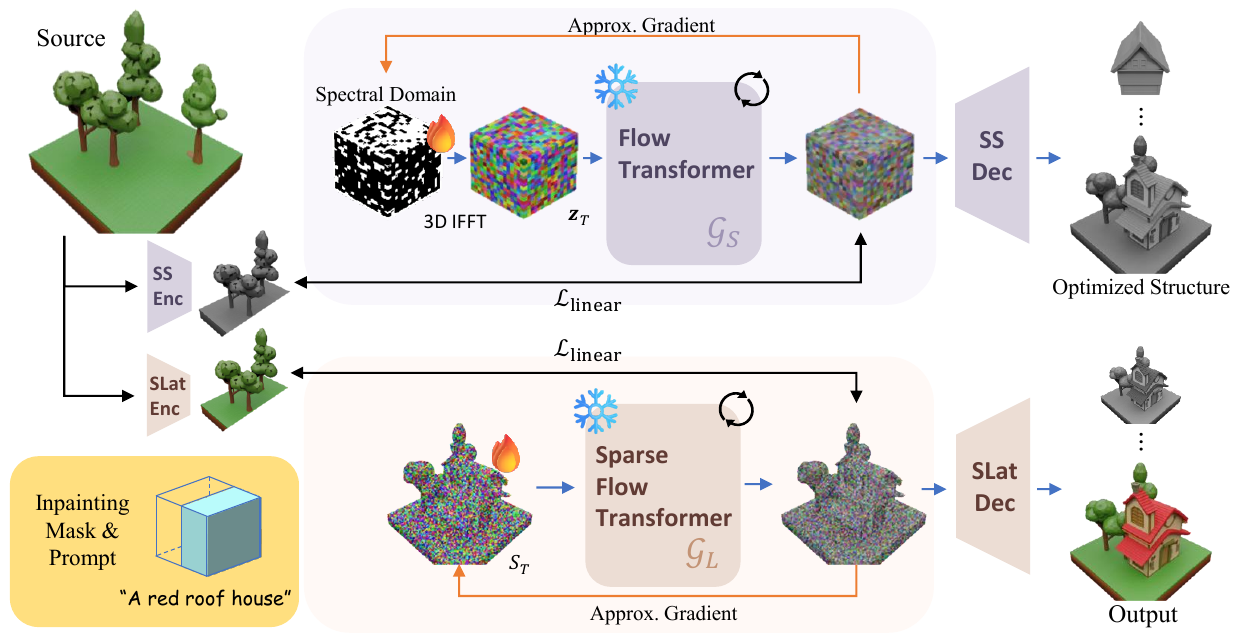}
    \caption{
        \textbf{Overview of InpaintSLAT.} By searching for an initial latent that preserves the conditioned region, our method generates results that satisfy the given prompt while maintaining the condition.
    }
    \label{fig:method}
    \vspace{-3mm}
\end{figure*}
%%%%%%%%%%%%%%%%%%%%%%%%%%%%%%%%%%%%%%%%%%%%%%5
\subsection{Optimize Initial Noisy Latent by Approximated Gradient} \label{sec:approx_gradient}
Let $x_T$ denote the initial noisy latent for the initial noisy sparse structure $S_T$ and initial noisy structured latent $\mathbf{z}_T$ of a rectified flow model $\mathcal{G}_S$ and $\mathcal{G}_L$, respectively.
A straightforward way to optimize $x_T$ would be to backpropagate through the entire denoising trajectory.
However, this requires storing intermediate states for all timesteps, which is memory-intensive and impractical for structured 3D latent generation. To enable efficient optimization, we approximate the denoising trajectory by locally linearizing it around the initial latent. The displacement-based formulation of rectified flow models in \cref{eq:RectifiedFlowModel} leads to locally smooth denoising dynamics, which allows effective first-order approximation of the trajectory. 
\begin{equation}
x(t) \approx x_T + (1 - \tfrac{t}{T})\,[D_T(x_T) - x_T]_{\mathrm{sg}},
\end{equation}
where $D_T(x_T)$ denote a single denoising prediction at timestep $T$ and $[\cdot]_{\mathrm{sg}}$ denotes the stop-gradient operator.
% The approximation uses the displacement direction $D_T(x_T)-x_T$ to enable efficient gradient computation with respect to the initial latent.
%
Under this linearized approximation for backpropagation, gradients can be propagated to the initial latent without storing the full trajectory. We therefore optimize the initial latent by minimizing
\begin{equation}
\mathcal{L}_{\text{linear}} =\big\|y - \mathcal{A}\!\left([D_T(x_T)-x_T]_{\mathrm{sg}} + x_T\right) \big\|_2^2 ,
\end{equation}
where $\mathcal{A}$ denotes the rendering or decoding operator and $y$ represents the target constraint.
By iteratively updating $x_T$ under this objective, we obtain an initialization that produces generations consistent with both the preserved structure and the target prompt.

Although the reverse diffusion dynamics are nonlinear, the displacement predicted by the denoiser provides a first-order estimate of the denoising direction. Consequently, the vector $D_T(x_T)-x_T$ serves as a useful proxy for the local trajectory direction, allowing us to construct an approximate linearization of the denoising process. Further, the optimization is performed iteratively via gradient descent, allowing the trajectory to be progressively corrected across iterations. Combined with the locally smooth dependence of the generation on the initial noise, this enables stable optimization of the initial latent.

%%%%%%%%%%%%%%%%%%%
\subsection{Optimization Strategies for SLAT-Conditioned Reconstruction} \label{sec:freq_param}
As observed in \cref{fig:optimization_process}, the sparse structure tends to change significantly during the early optimization stage ($t_{\mathrm{opt}}$), indicating that large updates are often required to steer the global geometry.
To accelerate convergence, we employ a relatively large learning rate during initialization optimization.
However, directly updating the voxel representation with a high learning rate often leads to unstable optimization.
To improve stability, we parameterize the sparse structure feature $S$ in the frequency domain by applying a 3D FFT to the voxel representation, inspired by Baek et al.~\cite{baek2025sonic}.
Optimization is then performed on the frequency coefficients rather than the voxel grid itself.
This representation allows global geometric adjustments to be captured by low-frequency components while suppressing high-frequency artifacts, leading to more stable optimization under large learning rates.

For structured latent features $\{z_i\}$, we instead perform direct optimization in the latent space.
Unlike the voxel-based sparse structure, these features are defined on irregular sparse coordinates and lack a regular grid structure, making frequency decomposition less suitable.
Moreover, the sparse structure already provides strong geometric constraints, as shown in \cref{fig:optimization_process}, which stabilize the optimization of $\{z_i\}$ in practice.

%%%%%%%%%%%%%%%%%%
\subsection{Optimization Strategies for Diversity and Quality} \label{sec:gauss_loss}
This initial latent optimization updates the latent to better reconstruct the preserved region, which causes the initial noise to deviate from the i.i.d. Gaussian distribution assumed by diffusion models. Interestingly, although the resulting noisy latent still reconstructs the preserved region well, the region generated according to the prompt becomes increasingly distorted as the number of optimization steps grows. To compensate for this distortion caused by overfitting, we introduce an additional loss that encourages the initial latent to remain close to a Gaussian distribution.
Let $\mu$, $\sigma$, skewness $\gamma$, and kurtosis $\kappa$ denote the first- to fourth-order statistics of the optimized latent $x_T$. We define the distribution regularization loss as
 
\begin{equation}
\mathcal{L}_{\text{dist}}=\lambda_1 \|\mu\|_2^2 +\lambda_2\|\sigma-1\|_2^2 +\lambda_3\|\gamma\|_2^2 +\lambda_4\|\kappa-3\|_2^2,
\end{equation}
which penalizes deviations from the statistics of $\mathcal{N}(0, \mathbf{I})$.
The final optimization objective is given by
\begin{equation}
\mathcal{L} =\mathcal{L}_{\text{recon}}+\mathcal{L}_{\text{dist}}.
\end{equation}
With this formulation, our method reliably reconstruct the preserved region while generating prompt-consistent content for the masked region. We validate the effectiveness of the proposed approach in the following experiments.
\section{Experiment} \label{sec:experiment}

\subsection{Experimental Setup}\

\subsubsection{Implementation Details}
We use the text-to-3D xlarge model of TRELLIS as the base generator. The diffusion process follows the default TRELLIS configuration with 12 sampling steps. For initial latent optimization, we perform $t_{opt}=15$ optimization steps using Adam optimizer with a learning rate of 5.0 for the sparse structure and 0.01 for the structured latent (SLAT). The hyperparameters for the Gaussian distribution regularization are set to $\lambda_1, \lambda_2, \lambda_3, \lambda_4 = 31.6, 10.0, 3.16, 1.0$, respectively. All experiments are conducted on NVIDIA A6000 GPUs. Additional implementation details are provided in the supplementary material.

\subsubsection{Datasets}
We conduct experiments on ABO~\cite{collins2022abo}, HSSD~\cite{khanna2024habitat}, and Toys4K~\cite{stojanov2021using}, which are part of the dataset collection used in TRELLIS. Since our method is training-free and uses TRELLIS as the backbone, we follow the TRELLIS evaluation protocol and report the main quantitative results on the Toys4K test set, while additional results on ABO and HSSD are provided in the supplementary material.

For evaluation, we consider the subset of Toys4K samples with captions (3180 out of 3229). For each sample, we construct a cubic mask where half of the volume is designated as the inpainting region and the other half as the preserved region. The model is conditioned on the preserved region and the caption, and generates geometry in the masked region consistent with both.

\begin{table}[tb]
  \caption{Quantitative Evaluation on Preserved Part in Toys4k Dataset.}
  \label{tab:recon_quan}
  \centering
  \resizebox{\linewidth}{!}{
  \begin{tabular}{@{}l|ccc|ccccc|ccc@{}}
    \toprule
    \multirow{3}{*}{\textbf{Method}} & \multicolumn{3}{c|}{\textbf{Appearance}} & \multicolumn{5}{c|}{\textbf{Geometry (Point Cloud)}} & \multicolumn{3}{c}{\textbf{Geometry (Normal Map)}} \\
    & \multirow{2}{*}{\textbf{PSNR$_\uparrow$}} & \multirow{2}{*}{\textbf{SSIM$_\uparrow$}} & \multirow{2}{*}{\textbf{LPIPS$_\downarrow$}} & \textbf{CD$_\downarrow$} & \textbf{Precision$_\uparrow$} & \textbf{Recall$_\uparrow$} & \textbf{F-score$_\uparrow$} & \textbf{F-score$_\uparrow$} & \multirow{2}{*}{\textbf{PSNR$_\uparrow$}} & \multirow{2}{*}{\textbf{SSIM$_\uparrow$}} & \multirow{2}{*}{\textbf{LPIPS$_\downarrow$}} \\
    & & & & L1{\tiny $\times100$} & @0.01 & @0.01 & @0.01 & @0.02 \\
    \midrule
    RePaint~\cite{lugmayr2022repaint}           & 15.14 & 0.702 & 0.2354 & 1.000 & 0.9028 & 0.9427 & 0.9191 & 0.9723 & 18.34 & 0.704 & 0.2110 \\
    SDEdit~\cite{meng2021sdedit}                & 15.63 & 0.717 & 0.2232 & 0.868 & 0.9216 & \underline{0.9658} & 0.9409 & 0.9820 & 19.36 & 0.726 & 0.1888 \\
    BlendedDiffusion~\cite{avrahami2022blended} & 15.13 & 0.702 & 0.2356 & 1.006 & 0.9017 & 0.9424 & 0.9182 & 0.9720 & 18.33 & 0.704 & 0.2111 \\
    MultiDiffusion~\cite{bar2023multidiffusion} & 15.06 & 0.701 & 0.2482 & 1.028 & 0.8993 & 0.9376 & 0.9147 & 0.9703 & 18.24 & 0.702 & 0.2136 \\
    DPS~\cite{chung2022diffusion}               & 14.65 & 0.694 & \underline{0.2160} & 1.183 & 0.8920 & 0.9163 & 0.8993 & 0.9609 & 17.67 & 0.692 & 0.2318 \\
    ILVR~\cite{choi2021ilvr}                    & \underline{16.00} & \underline{0.724} & 0.2370 & \underline{0.823} & \underline{0.9279} & \textbf{0.9743} & \underline{0.9485} & \underline{0.9856} & \underline{19.76} & \underline{0.736} & \underline{0.1779} \\
    Ours                                        & \textbf{18.56} & \textbf{0.793} & \textbf{0.1455} & \textbf{0.637} & \textbf{0.9483} & \textbf{0.9681} & \textbf{0.9577} & \textbf{0.9922} & \textbf{22.26} & \textbf{0.831} & \textbf{0.0943} \\
  \bottomrule
  \end{tabular}
  }
\end{table}

\begin{table}[tb]
  \caption{Quantitative Evaluation on Inpainting Part in Toys4k Dataset.}
  \label{tab:gen_quan}
  \centering
  \resizebox{0.6\linewidth}{!}{
  \begin{tabular}{@{}l|ccccc|c@{}}
    \toprule
    Method & CLIP$_\uparrow$ & FD$_{dinov2j,\downarrow}$ & KD$_{dinov2,\downarrow}$ & Runtime(s)$_\downarrow$ \\
    \midrule
    RePaint~\cite{lugmayr2022repaint}           & 30.17 & 19.56 & 1.377 & 24.76 \\
    SDEdit~\cite{meng2021sdedit}                & 29.16 & 17.56 & 1.125 & 28.34 \\
    BlendedDiffusion~\cite{avrahami2022blended} & 30.17 & 19.54 & 1.362 & \underline{24.68} \\
    MultiDiffusion~\cite{bar2023multidiffusion} & 30.18 & 19.69 & 1.381 & 29.92 \\
    DPS~\cite{chung2022diffusion}               & 30.01 & 22.67 & 1.601 & \textbf{22.96} \\
    ILVR~\cite{choi2021ilvr}                    & \textbf{30.61} & \underline{12.09} & \underline{0.758} & 29.24 \\
    Ours                                        & \underline{30.42} & \textbf{3.683} & \textbf{0.194} & 145.2 \\
  \bottomrule
  \end{tabular}
  }
\end{table}

\begin{figure*}[t]
    \includegraphics[width=1.0\textwidth]{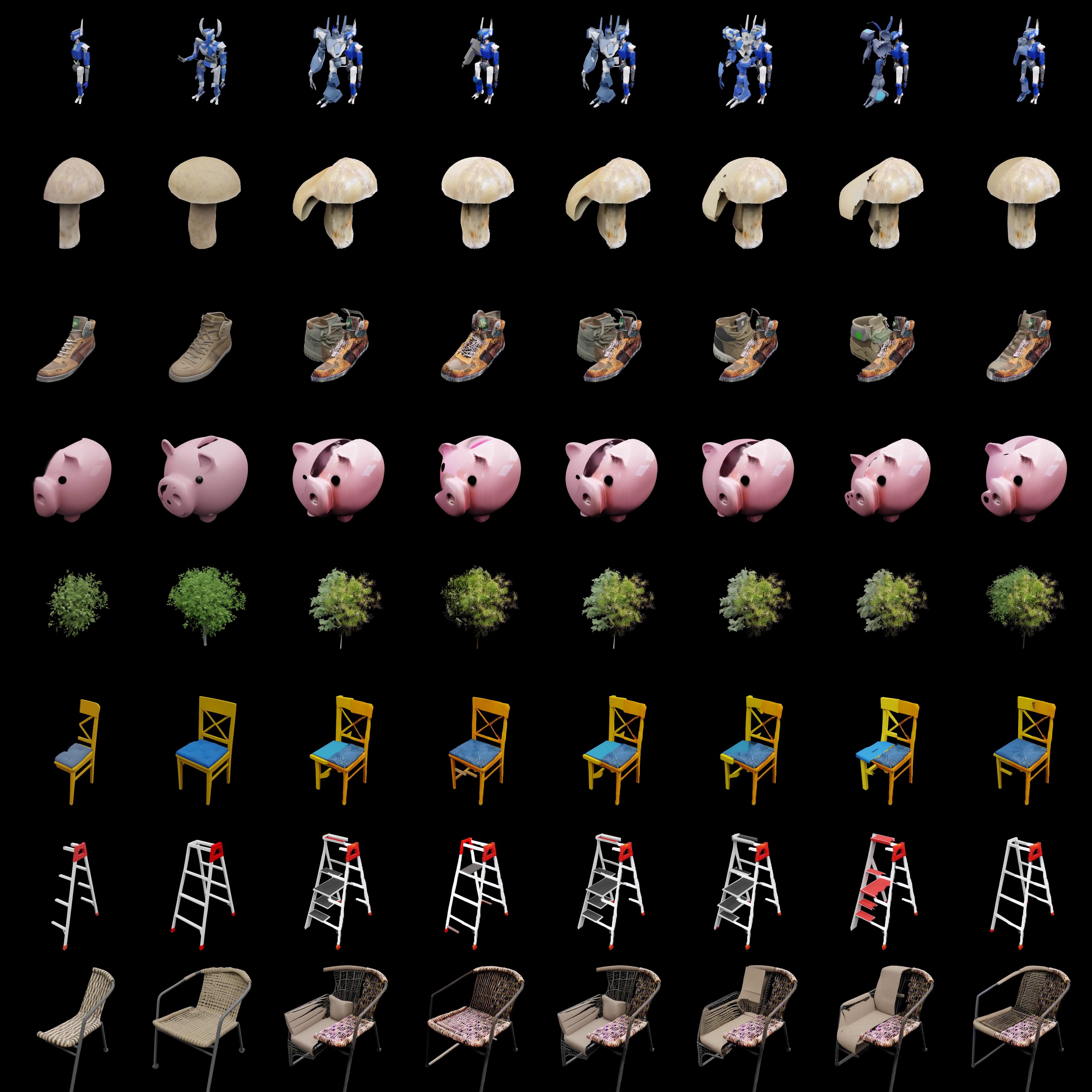}
    \vspace{-6mm}
    \caption{
        Qualitative Results for Toys4k
    }
    \label{fig:qual_app}
    \vspace{-6mm}
\end{figure*}
\begin{figure*}[t]
    \includegraphics[width=1.0\textwidth]{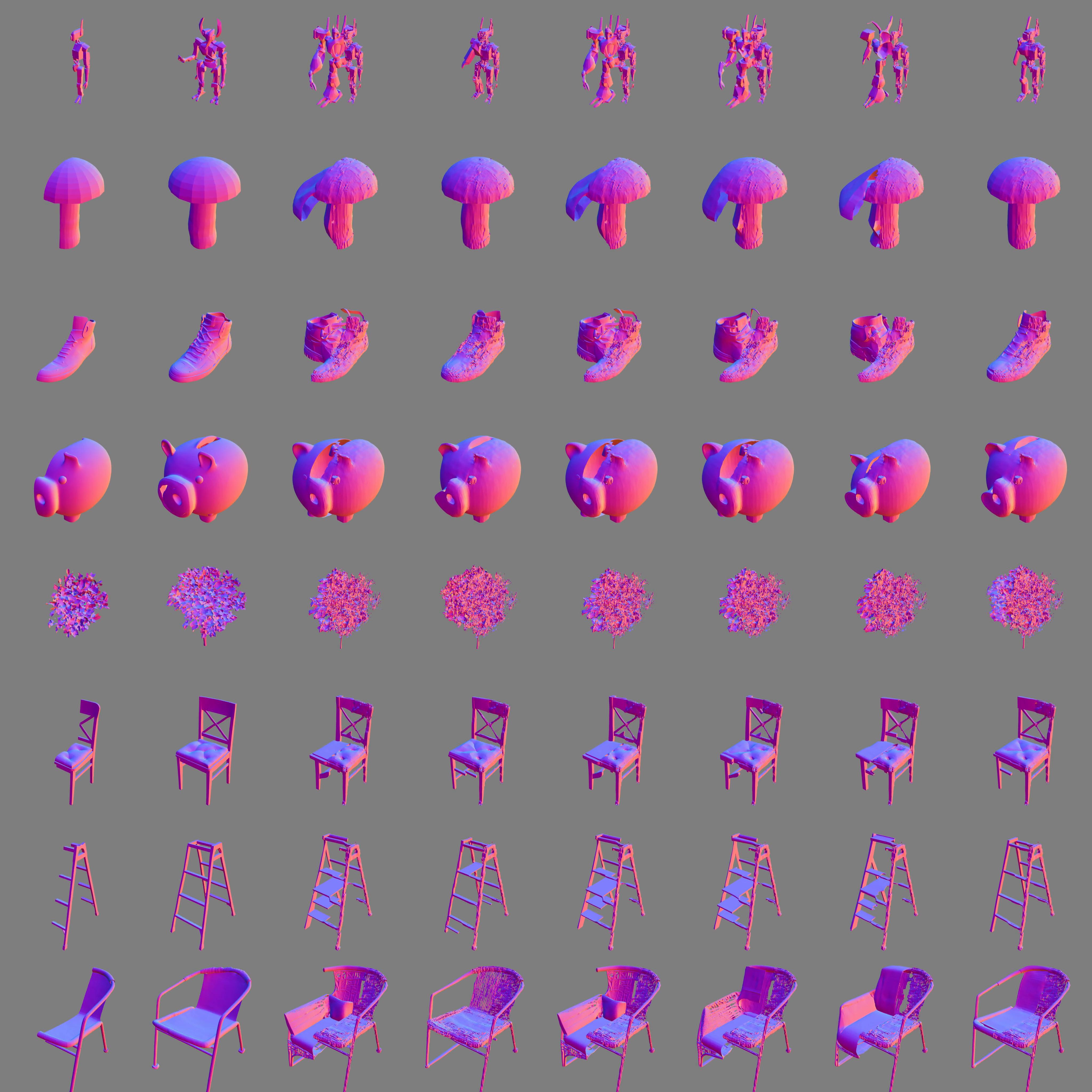}
    \vspace{-6mm}
    \caption{
        Geometry samples Qualitative Results for Toys4k
    }
    \label{fig:qual_geo}
    \vspace{-6mm}
\end{figure*}

\subsection{Baseline}
We compare our method with several representative diffusion-based editing and conditional generation approaches, including RePaint, SDEdit, MultiDiffusion, Blended Diffusion, DPS, and ILVR. These methods cover a broad range of diffusion-based editing strategies such as inpainting, noise-based editing, spatially constrained generation, gradient-based editing, and posterior-guided sampling.
RePaint~\cite{lugmayr2022repaint} is a diffusion-based inpainting method that repeatedly resamples masked regions while enforcing consistency with known regions. It serves as a strong baseline for diffusion-based inpainting.
SDEdit~\cite{meng2021sdedit} performs editing by adding noise to the input and subsequently denoising it with a diffusion model. It represents a simple yet effective approach for structure-preserving diffusion editing.
Blended Diffusion~\cite{avrahami2022blended} performs localized editing by blending gradients from the conditioned region with diffusion updates, enabling spatially restricted modifications.
MultiDiffusion~\cite{bar2023multidiffusion} enables spatially constrained generation by combining diffusion predictions from multiple overlapping windows, allowing region-level control during generation.
DPS~\cite{chung2022diffusion} (Diffusion Posterior Sampling) guides diffusion sampling using a likelihood-based posterior update, enabling the generation process to satisfy external constraints.
ILVR~\cite{choi2021ilvr} guides diffusion generation by iteratively enforcing consistency with a reference signal in the low-frequency domain.

\subsubsection{Evaluation metrics}
We evaluate our method from two perspectives: (1) how well the preserved region is reconstructed and (2) how well the inpainting region is generated according to the text prompt. 
The reconstruction results are evaluated in terms of both appearance and geometry. For geometry evaluation, we separately assess point clouds and surface normals. Point clouds are evaluated using Chamfer Distance (L1) and F-score, while surface normals are evaluated using image-based metrics computed from normal map renderings, denoted as PSNR-N (Normal), SSIM-N, and LPIPS-N.
To evaluate the quality of the content generated in the inpainting region, we computed the CLIP~\cite{radford2021learning} score between the results and the captions provided in Toys4K. Additionally, to measure the similarity of the generated sample distribution, we extract DINOv2~\cite{oquab2023dinov2} features and compute Fréchet Distance (FD)~\cite{heusel2017gans} and Kernel Distance (KD)~\cite{binkowski2018demystifying}.

\begin{figure*}[t]
    \includegraphics[width=1.0\textwidth]{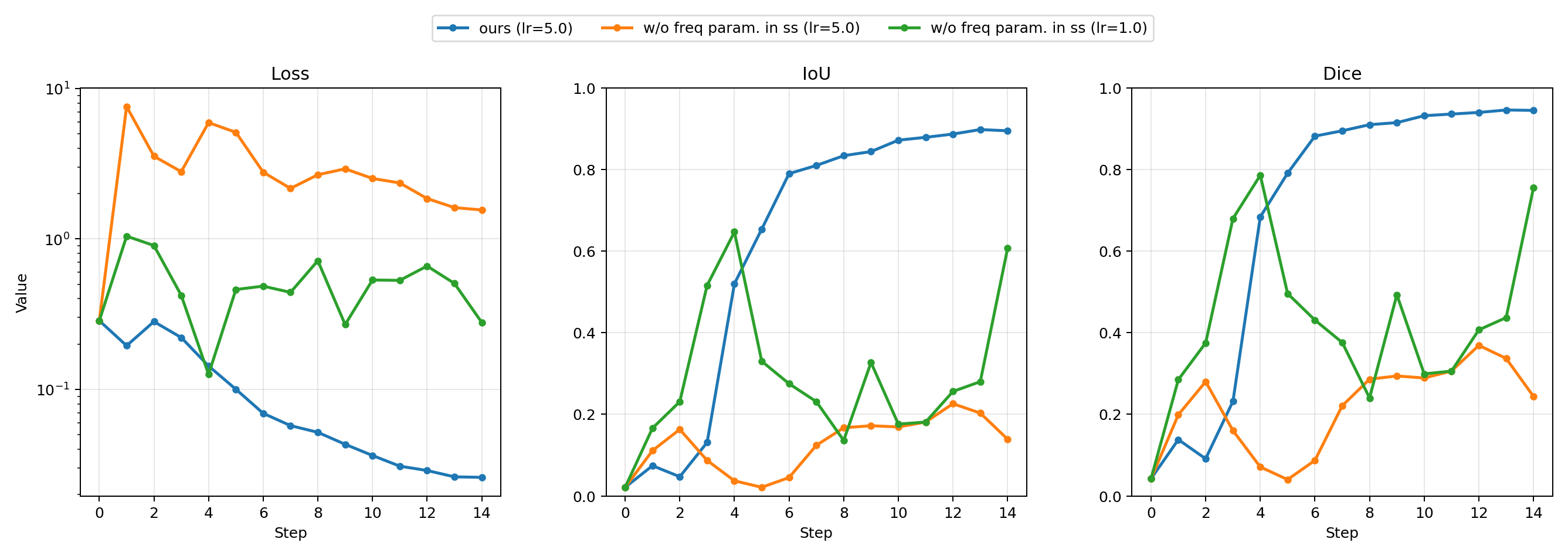}
    \caption{
        Optimization stability of the frequency-domain parameterization for sparse structure.
    }
    \label{fig:freq_param}
    \vspace{-3mm}
\end{figure*}
\begin{figure*}[t]
    \includegraphics[width=1.0\textwidth]{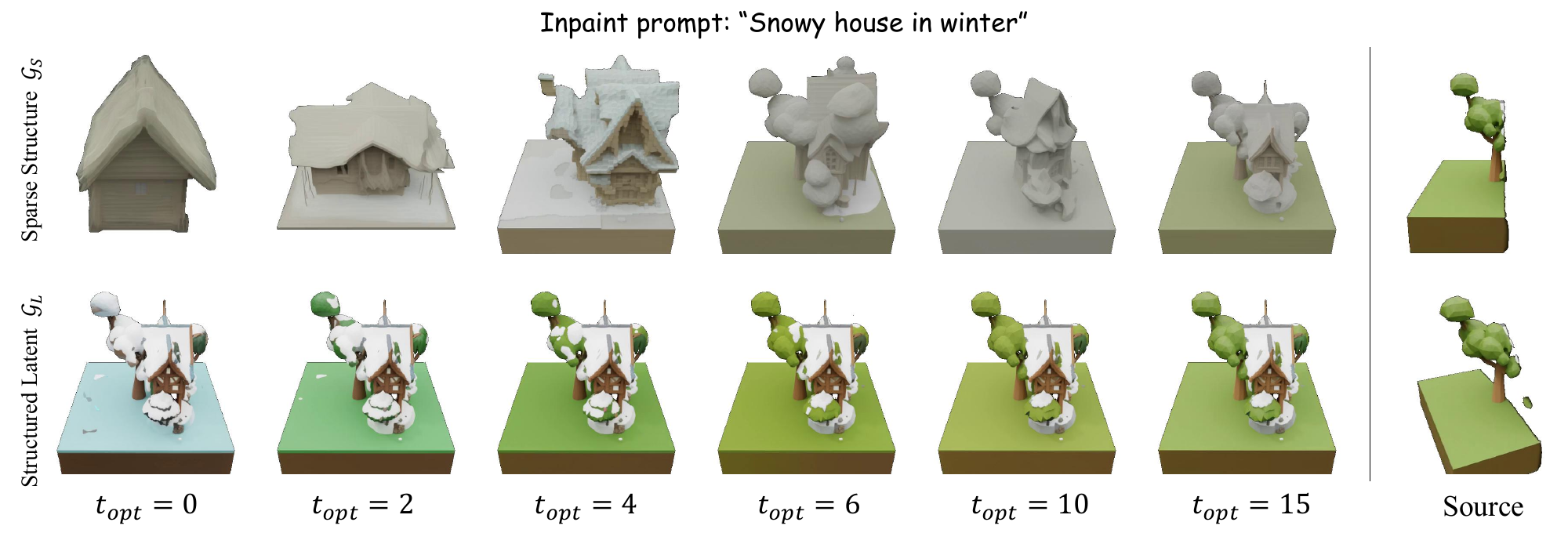}
    \caption{
        Optimization process of sparse structure and structured latent.
    }
    \label{fig:optimization_process}
    \vspace{-3mm}
\end{figure*}

\subsection{Inpainting result}
We report quantitative results for reconstruction and generation in \cref{tab:recon_quan} and \cref{tab:gen_quan}, respectively. Qualitative comparisons of appearance and normal map renderings are shown in \cref{fig:qual_app} and \cref{fig:qual_geo}.
Overall, our method outperforms existing baselines not only in appearance quality  but also in geometric fidelity evaluated using point clouds and surface normals. 
One notable limitation of our approach is runtime for searching a suitable initial latent through iterative optimization. It requires multiple iterations and thus results in a relatively longer runtime. 
We report runtime using a fixed configuration of $t_{opt}=15$ optimization steps for stable evaluation across all samples. In practice, however, the optimization converges much earlier. As shown in \cref{fig:freq_param}, IoU exceeds 0.8 and Dice exceeds 0.9 within only six optimization steps (blue graph). This demonstrates a favorable trade-off between generation quality and runtime.
This convergence behavior around $t_{opt}=6$ is also visually evident in \cref{fig:optimization_process}. The sparse structure, which models coarse geometry, and the structured latent, which captures appearance and fine details, are already close to their final states at this stage.

\begin{table}[tb]
  \caption{Quantitative Evaluation on Preserved Part with Reconstruction metrics (Toys4k).}
  \label{tab:ablation}
  \centering
  \resizebox{\linewidth}{!}{
  \begin{tabular}{@{}l|ccc|ccccc|ccc@{}}
    \toprule
    \multirow{3}{*}{\textbf{Method}} & \multicolumn{3}{c|}{\textbf{Appearance}} & \multicolumn{5}{c|}{\textbf{Geometry (Point Cloud)}} & \multicolumn{3}{c}{\textbf{Geometry (Normal Map)}} \\
    & \multirow{2}{*}{\textbf{PSNR$_\uparrow$}} & \multirow{2}{*}{\textbf{SSIM$_\uparrow$}} & \multirow{2}{*}{\textbf{LPIPS$_\downarrow$}} & \textbf{CD$_\downarrow$} & \textbf{Precision$_\uparrow$} & \textbf{Recall$_\uparrow$} & \textbf{F-score$_\uparrow$} & \textbf{F-score$_\uparrow$} & \multirow{2}{*}{\textbf{PSNR$_\uparrow$}} & \multirow{2}{*}{\textbf{SSIM$_\uparrow$}} & \multirow{2}{*}{\textbf{LPIPS$_\downarrow$}} \\
    & & & & L1{\tiny $\times100$} & @0.01 & @0.01 & @0.01 & @0.02 \\
    \midrule
    Ours & \textbf{18.56} & \textbf{0.793} & \textbf{0.1455} & \textbf{0.637} & \textbf{0.9483} & \textbf{0.9681} & \textbf{0.9577} & \textbf{0.9922} & \textbf{22.26} & \textbf{0.831} & \textbf{0.0943} \\
    w/o Freq. Param. (\cref{sec:freq_param}) & 9.64 & 0.475 & 0.1454 & 14.42 & 0.2011 & 0.4184 & 0.2402 & 0.3961 & 13.08 & 0.518 & 0.5443 \\
    w/o Gaussian Loss (\cref{sec:gauss_loss}) & 17.91 & 0.792 & 0.5413 & 0.691 & 0.9433 & 0.9645 & 0.9533 & 0.9910 & 21.62 & 0.8236 & 0.0967 \\
    
  \bottomrule
  \end{tabular}
  }
\end{table}

\begin{figure*}[t]
    \includegraphics[width=1.0\textwidth]{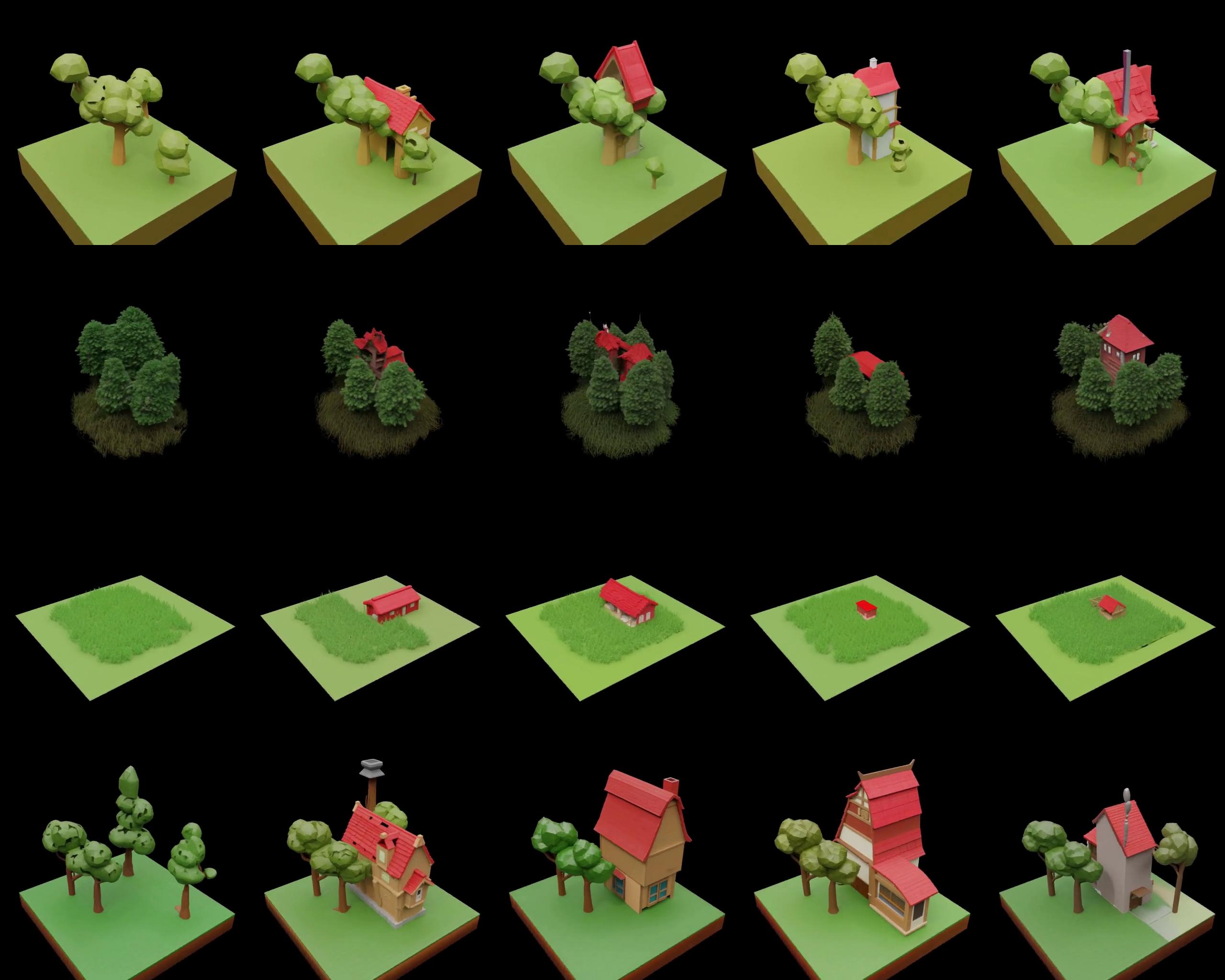}
    \caption{
        Qualitative results with manual prompt and various seeds.
    }
    \label{fig:various_seeds}
    \vspace{-3mm}
\end{figure*}

\begin{figure*}[t]
    \centering
    \includegraphics[width=0.90\textwidth]{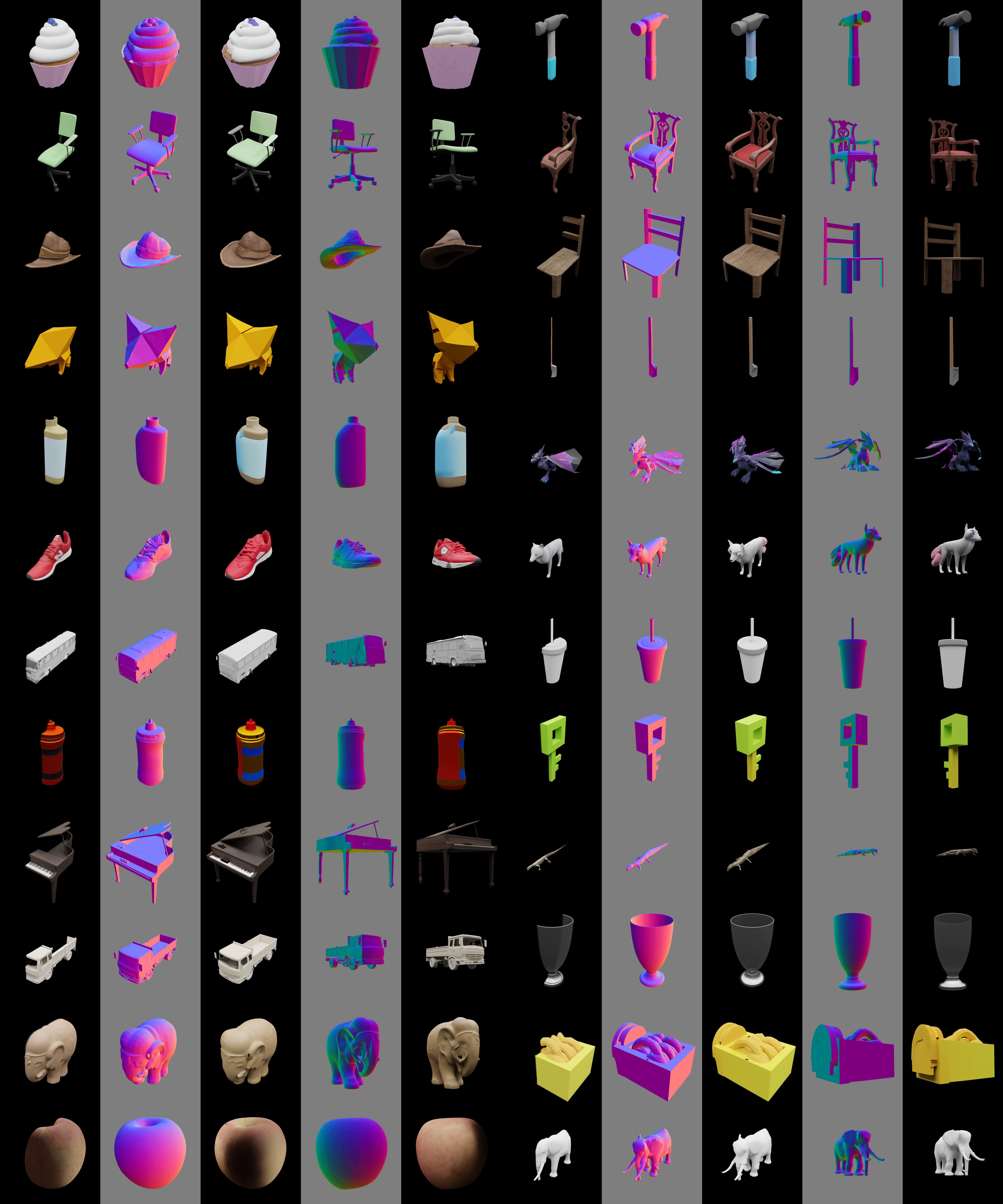}
    \caption{
        Qualitative results for inpainting in Toys4k~\cite{stojanov2021using}
    }
    \label{fig:toys4k_show}
    \vspace{-6mm}
\end{figure*}

%%%%%%%%%%%%%%%%%%%%%%
\subsection{Ablation Study}
We conduct ablation studies on the proposed spectral-domain parameterization and the Gaussian distribution matching loss, and report the results in \cref{tab:ablation}. Optimizing the structured latent without spectral-domain parameterization leads to severe performance degradation. As shown in \cref{fig:freq_param}, using a lower learning rate appears to stabilize the optimization to some extent, but it still fails to achieve proper convergence. In contrast, with spectral-domain parameterization, our method can employ a higher learning rate and converges rapidly within only a few optimization steps.

We also evaluate the effect of the Gaussian distribution matching loss, which encourages the initial latent to remain close to a Gaussian distribution. As shown in \cref{tab:ablation}, removing this loss results in unstable inpainting behavior, whereas incorporating it enables more stable and reliable inpainting.
\cref{fig:various_seeds} shows generation results using the same prompt, \textit{“A red roof house,”} with different random seeds. The results demonstrate that our method consistently preserves the conditioned region while producing diverse inpainting outcomes across different seeds.
For reference, we visualize the inpainting result with Toys4k dataset in \cref{fig:toys4k_show}.

\section{Conclusion and Limitation}
We presented a training-free approach for 3D inpainting that operates by optimizing the initial structured latent seed of a pretrained 3D diffusion model. Rather than modifying model parameters or manipulating the sampling trajectory, our method refines the initialization under contextual reconstruction constraints, allowing the pretrained generative prior to synthesize masked regions while preserving geometric consistency in unmasked areas. By combining memory-efficient approximate backpropagation, structured basis parameterization, and reduced-step optimization, we achieve stable and efficient control over 3D generation.

Our approach nevertheless inherits a fundamental limitation of prior-based generative modeling. Because we search for a suitable initialization within the learned distribution of the pretrained model, successful inpainting depends on the compatibility between the preserved structure and the target prompt. When the contextual geometry and textual condition are highly inconsistent or underrepresented in the training data, the model may fail to synthesize plausible completions. Addressing such prior limitations, potentially through stronger structural conditioning or hybrid guidance mechanisms, remains an important direction for future research.
% \section*{Acknowledgements}
% Please insert your acknowledgments here.

\bibliographystyle{splncs04}
\bibliography{main}

\newpage
\appendix
\clearpage
\setcounter{page}{1}
% \maketitlesupplementary

\setcounter{section}{0}
\setcounter{figure}{0}
\setcounter{table}{0}

\renewcommand{\thesection}{S\arabic{section}}
\renewcommand{\thetable}{S\arabic{table}}
\renewcommand{\thefigure}{S\arabic{figure}}

\title{\textit{Supplementary Materials for} \\ InpaintSLat: Inpainting Structured 3D Latents via Noise Optimization
} 
\author{}
\institute{}
\maketitle

%% ABO, HSSD의 (1) recon/gen table, (2) app/geo qual
%% opt step ablation
%% Video
%% Limitation

\section{Additional Experiment Results}
As mentioned in the main text, we additionally report experimental results on the ABO and HSSD datasets. Since both datasets were used during the training of TRELLIS, we follow the same data generation process as TRELLIS, which makes the relative performance improvement of our method more evident. As shown in \cref{tab:abo_recon_quan} and \cref{tab:hssd_recon_quan}, which report reconstruction metrics, existing methods achieve comparable performance on the ABO and HSSD datasets, whereas our method consistently reports higher scores in both appearance and geometry reconstruction.

\cref{tab:abo_gen_quan} and \cref{tab:hssd_gen_quan} present the generation metrics, where the differences between methods appear to be relatively small. Although FD and KD measure distributional similarity and are therefore not straightforward to compare directly, we believe that the limited number of samples is the primary reason for the lack of clear differences. In our evaluation, the metrics were computed using one randomly selected view image per 3D object. For a more reliable assessment, it would be preferable to evaluate the metrics using samples generated from multiple random seeds for each method.

\begin{figure*}[t]
    \includegraphics[width=1.0\textwidth]{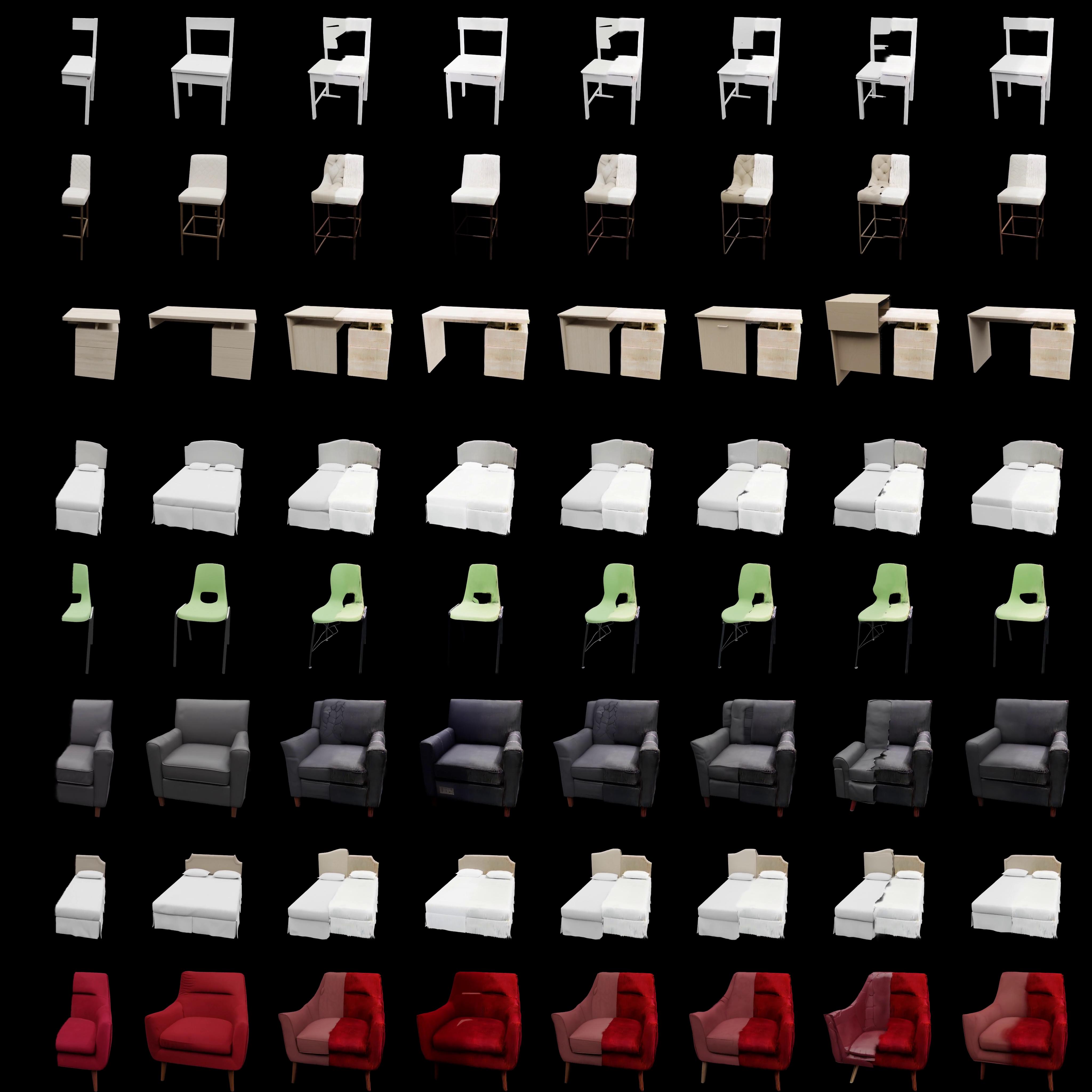}
    \caption{
        Qualitative Results for ABO
    }
    \label{fig:supp_abo_qual_app}
    \vspace{-6mm}
\end{figure*}

\begin{figure*}[t]
    \includegraphics[width=1.0\textwidth]{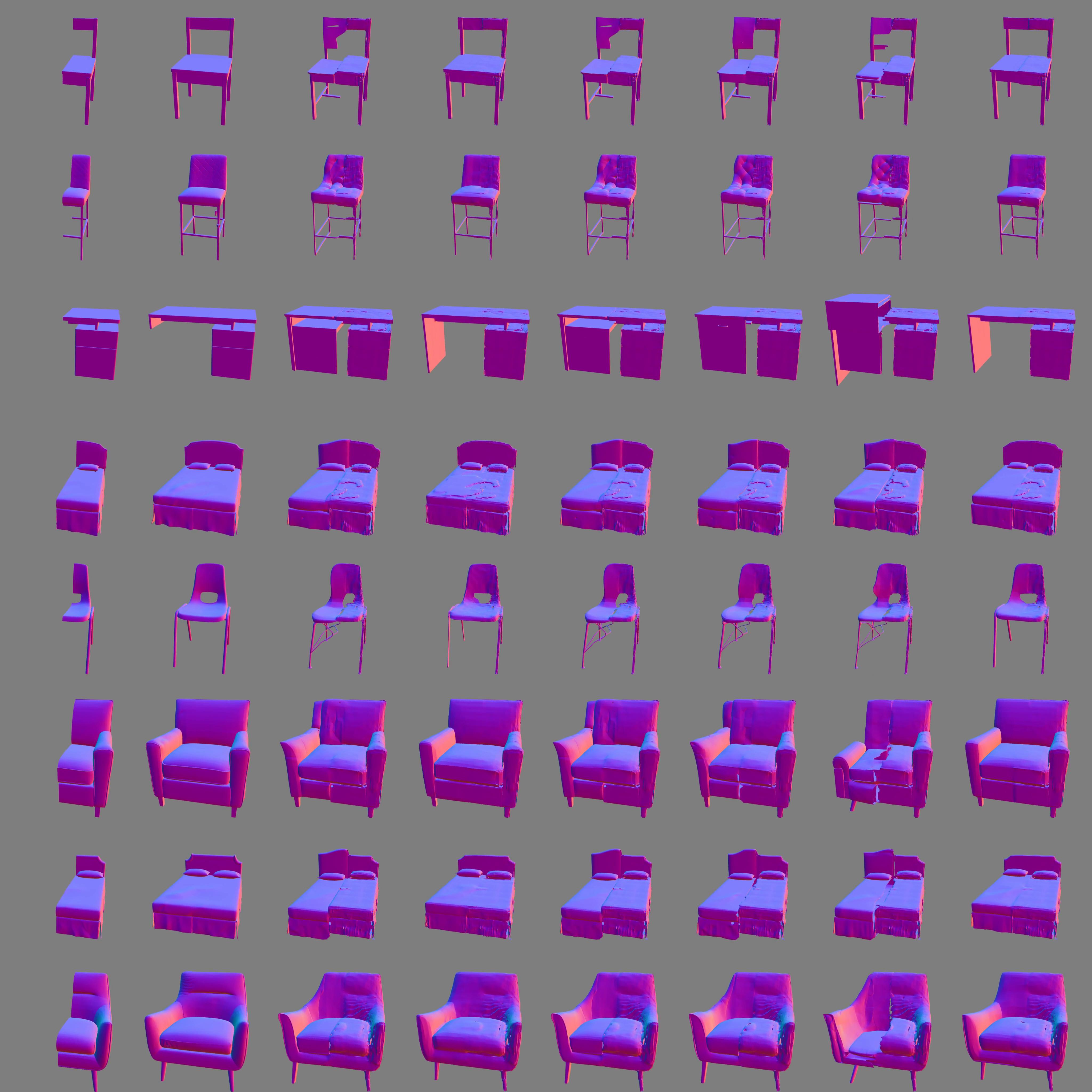}
    \caption{
        Qualitative Results for ABO
    }
    \label{fig:supp_abo_qual_geo}
    \vspace{-6mm}
\end{figure*}

\begin{figure*}[t]
    \includegraphics[width=1.0\textwidth]{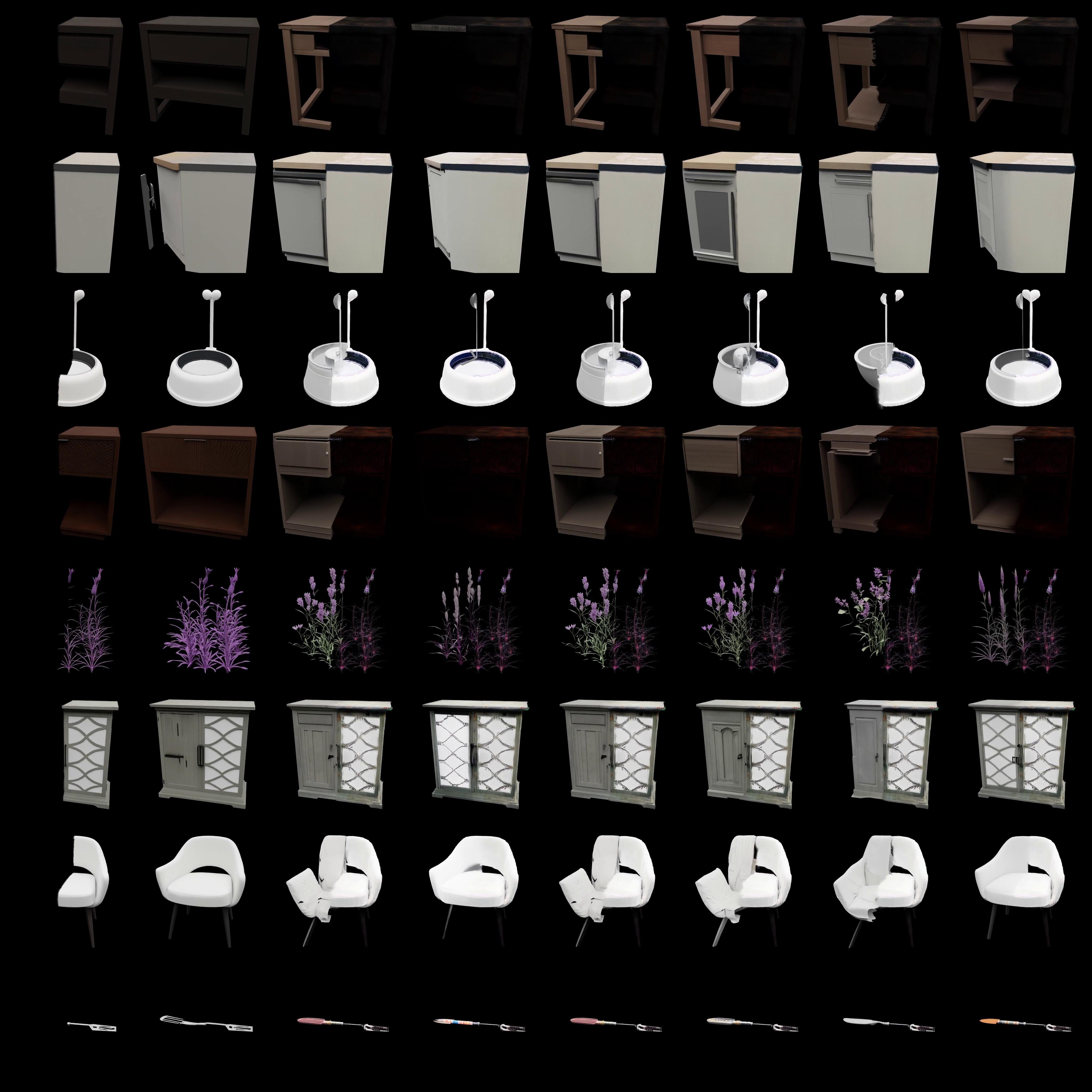}
    \caption{
        Qualitative Results for HSSD
    }
    \label{fig:supp_hssd_qual_app}
    \vspace{-6mm}
\end{figure*}

\begin{figure*}[t]
    \includegraphics[width=1.0\textwidth]{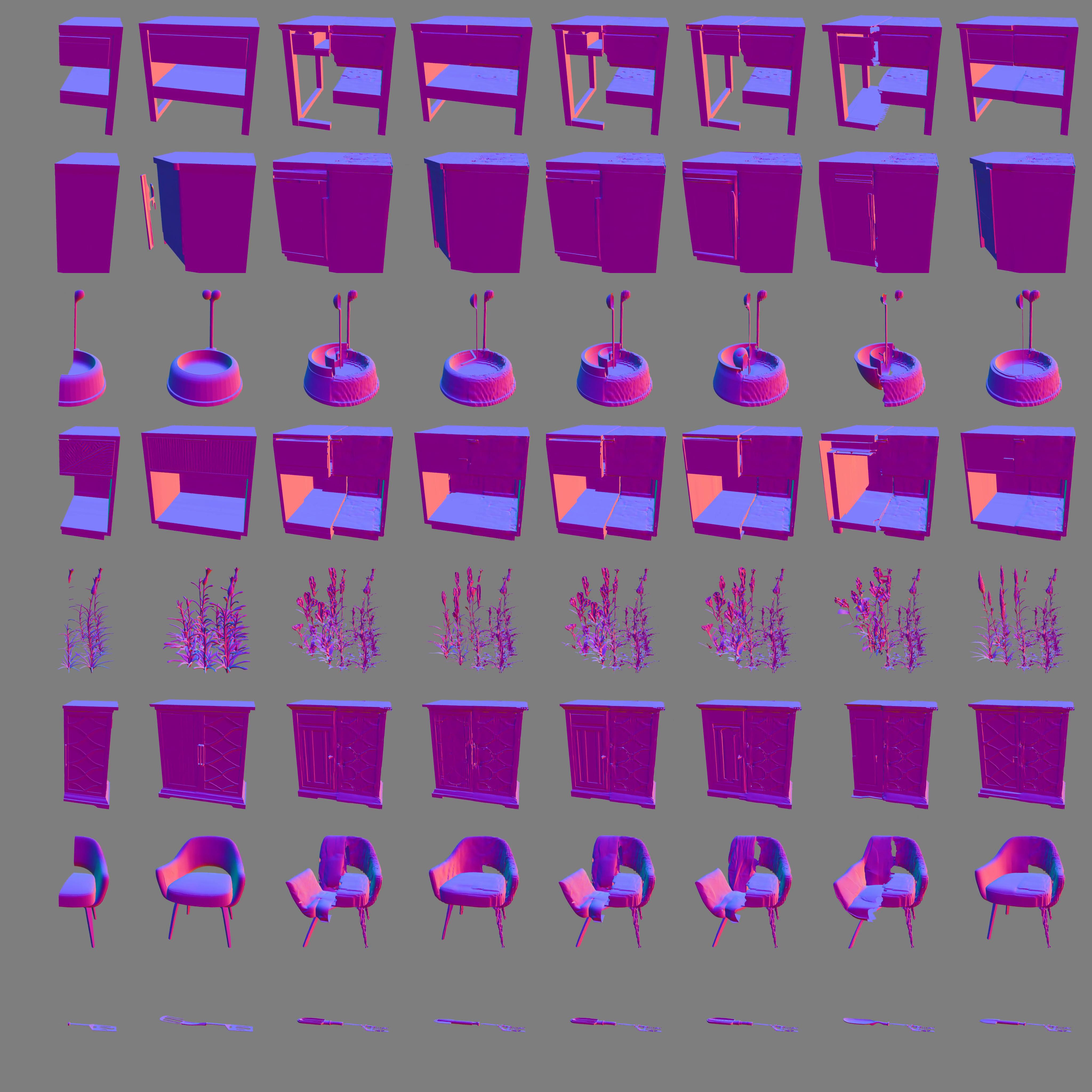}
    \caption{
        Qualitative Results for HSSD
    }
    \label{fig:supp_hssd_qual_geo}
    \vspace{-6mm}
\end{figure*}

To qualitatively assess the results, we present the appearance in \cref{fig:supp_abo_qual_app} and \cref{fig:supp_hssd_qual_app} and the corresponding geometry renderings in \cref{fig:supp_abo_qual_geo} and \cref{fig:supp_hssd_qual_geo}. We also provide supplementary videos showing results generated on the Toys4K, ABO, and HSSD datasets for further reference.

% \section{Ablation on optimization steps}
% As mentioned in the main text, our proposed method searches for an initial noise that satisfies the given conditions by repeating the generation process multiple times. However, this repeated generation procedure incurs substantial computational and time costs.

% To ensure robust and consistent reporting of results, we use the default diffusion setting of TRELLIS (12 steps) and present results obtained with $t_{opt}=15$ optimization iterations. In practice, the method can be operated more efficiently by reducing the number of diffusion steps or adopting an adaptive strategy for these steps, which ultimately results in a performance–time trade-off. We report the results of experiments with varying diffusion steps and optimization steps $t_{opt}=15$ in Table 5.

\section{Discussion}
% * simple gradient를 위한 approx (eq 3) rectified flow 위에서도 강한 가정이다.
% Approximated gradient는 대략적인 방향만 맞는데, gradient descent가 금방 데러다 줌 + latent manifold가 local smooth해서 근처까지만 가도 비슷하게 나옴.
% -> fine detail은 잘 안 됌 (displacement 잧가 low-freq) + 정확히 일치하지 않음
% 6 iter도 잘 됌 (6 iter면 근처 smooth까진 오고, 이후 opt는 좀 잘해야 겠는데)
% 
Equation 3 is introduced to enable a simplified backpropagation computation. However, this assumption is quite strong, even for TRELLIS, which is based on rectified flow. As a result, the approximated gradient only captures the general direction of the update rather than the exact gradient.

Despite this limitation, gradient descent can quickly move the solution toward a nearby plausible region in the latent space. Once the latent reaches this neighborhood, the subsequent diffusion process tends to produce similar outputs. We speculate that this behavior arises from the locally smooth structure of the latent manifold. Because of this property, reaching a nearby point is often sufficient to generate comparable results, even if the optimization does not converge to the exact optimum.

% \paragraph{Limitations and Future work}
% 더 많은 data로 학습해보기 (굳이 scene data가 아니어도 됨)

\paragraph{Social Impact.} Our work focuses on controllable 3D scene generation and does not directly address personal data, identity modeling, or downstream decision-making tasks. As such, we do not anticipate significant negative social impacts. Potential applications—such as simulation, virtual environment creation, and content generation—are primarily creative or industrial, and the method does not inherently facilitate misuse beyond the general considerations associated with generative models. We encourage responsible use within appropriate ethical and safety guidelines.

\begin{table}[tb]
  \caption{Quantitative Evaluation on Preserved Part with Reconstruction metrics (ABO).}
  \label{tab:abo_recon_quan}
  \centering
  \resizebox{\linewidth}{!}{
  \begin{tabular}{@{}l|ccc|ccccc|ccc@{}}
    \toprule
    \multirow{3}{*}{\textbf{Method}} & \multicolumn{3}{c|}{\textbf{Appearance}} & \multicolumn{5}{c|}{\textbf{Geometry (Point Cloud)}} & \multicolumn{3}{c}{\textbf{Geometry (Normal Map)}} \\
    & \multirow{2}{*}{\textbf{PSNR$_\uparrow$}} & \multirow{2}{*}{\textbf{SSIM$_\uparrow$}} & \multirow{2}{*}{\textbf{LPIPS$_\downarrow$}} & \textbf{CD$_\downarrow$} & \textbf{Precision$_\uparrow$} & \textbf{Recall$_\uparrow$} & \textbf{F-score$_\uparrow$} & \textbf{F-score$_\uparrow$} & \multirow{2}{*}{\textbf{PSNR$_\uparrow$}} & \multirow{2}{*}{\textbf{SSIM$_\uparrow$}} & \multirow{2}{*}{\textbf{LPIPS$_\downarrow$}} \\
    & & & & L1{\tiny $\times100$} & @0.01 & @0.01 & @0.01 & @0.02 \\
    \midrule
    RePaint~\cite{lugmayr2022repaint}           & 16.83 & 0.753 & 0.2132 & 1.238 & 0.8831 & 0.9688 & 0.9175 & 0.9620 & 20.51 & 0.788 & 0.1471 \\
    SDEdit~\cite{meng2021sdedit}                & 16.97 & 0.760 & 0.2059 & 1.018 & 0.9300 & 0.9727 & 0.9460 & 0.9774 & 21.15 & 0.801 & 0.1342 \\
    BlendedDiffusion~\cite{avrahami2022blended} & 16.84 & 0.753 & 0.2135 & 1.244 & 0.8833 & 0.9686 & 0.9174 & 0.9618 & 20.51 & 0.788 & 0.1473 \\
    MultiDiffusion~\cite{bar2023multidiffusion} & 16.81 & 0.752 & 0.2154 & 1.307 & 0.8754 & 0.9679 & 0.9125 & 0.9585 & 20.41 & 0.786 & 0.1501 \\
    DPS~\cite{chung2022diffusion}               & 16.41 & 0.745 & 0.2270 & 1.550 & 0.8290 & 0.9604 & 0.8823 & 0.9382 & 19.68 & 0.774 & 0.1691 \\
    ILVR~\cite{choi2021ilvr}                    & 17.38 & 0.767 & 0.1993 & 0.965 & 0.9406 & 0.9751 & 0.9534 & 0.9820 & 21.45 & 0.807 & 0.1291 \\
    Ours                                        & 25.14 & 0.908 & 0.0650 & 0.541 & 0.9883 & 0.9883 & 0.9876 & 0.9953 & 27.3252 & 0.9227 & 0.0352 \\
  \bottomrule
  \end{tabular}
  }
\end{table}

\begin{table}[tb]
  \caption{Quantitative Evaluation on Inpainting Part with Generation metrics (ABO).}
  \label{tab:abo_gen_quan}
  \centering
  \resizebox{0.7\linewidth}{!}{
  \begin{tabular}{@{}l|ccccc|c@{}}
    \toprule
    Method & CLIP$_\uparrow$ & FD$_{dinov2j,\downarrow}$ & KD$_{dinov2,\downarrow}$ \\
    \midrule
    RePaint~\cite{lugmayr2022repaint}           & 28.15 & 26.44 & 2.701 \\
    SDEdit~\cite{meng2021sdedit}                & 26.90 & 19.17 & 1.497 \\
    BlendedDiffusion~\cite{avrahami2022blended} & 28.15 & 26.51 & 2.633 \\
    MultiDiffusion~\cite{bar2023multidiffusion} & 28.09 & 27.17 & 2.641 \\
    DPS~\cite{chung2022diffusion}               & 27.80 & 32.86 & 3.095 \\
    ILVR~\cite{choi2021ilvr}                    & 28.25 & 15.97 & 1.248 \\
    Ours                                        & 28.36 & 9.238 & 0.570 \\
  \bottomrule
  \end{tabular}
  }
\end{table}

\begin{table}[tb]
  \caption{Quantitative Evaluation on Preserved Part with Reconstruction metrics (Toys4k).}
  \label{tab:hssd_recon_quan}
  \centering
  \resizebox{\linewidth}{!}{
  \begin{tabular}{@{}l|ccc|ccccc|ccc@{}}
    \toprule
    \multirow{3}{*}{\textbf{Method}} & \multicolumn{3}{c|}{\textbf{Appearance}} & \multicolumn{5}{c|}{\textbf{Geometry (Point Cloud)}} & \multicolumn{3}{c}{\textbf{Geometry (Normal Map)}} \\
    & \multirow{2}{*}{\textbf{PSNR$_\uparrow$}} & \multirow{2}{*}{\textbf{SSIM$_\uparrow$}} & \multirow{2}{*}{\textbf{LPIPS$_\downarrow$}} & \textbf{CD$_\downarrow$} & \textbf{Precision$_\uparrow$} & \textbf{Recall$_\uparrow$} & \textbf{F-score$_\uparrow$} & \textbf{F-score$_\uparrow$} & \multirow{2}{*}{\textbf{PSNR$_\uparrow$}} & \multirow{2}{*}{\textbf{SSIM$_\uparrow$}} & \multirow{2}{*}{\textbf{LPIPS$_\downarrow$}} \\
    & & & & L1{\tiny $\times100$} & @0.01 & @0.01 & @0.01 & @0.02 \\
    \midrule
    RePaint~\cite{lugmayr2022repaint}           & 16.01 & 0.690 & 0.2496 & 1.081 & 0.8945 & 0.9569 & 0.9186 & 0.9668 & 19.54 & 0.742 & 0.1766 \\
    SDEdit~\cite{meng2021sdedit}                & 16.19 & 0.697 & 0.2420 & 0.928 & 0.9327 & 0.9665 & 0.9458 & 0.9807 & 20.33 & 0.756 & 0.1615 \\
    BlendedDiffusion~\cite{avrahami2022blended} & 16.02 & 0.690 & 0.2497 & 1.085 & 0.8910 & 0.9578 & 0.9168 & 0.9662 & 19.56 & 0.742 & 0.1763 \\
    MultiDiffusion~\cite{bar2023multidiffusion} & 15.99 & 0.689 & 0.2510 & 1.101 & 0.8845 & 0.9579 & 0.9132 & 0.9646 & 19.53 & 0.740 & 0.1786 \\
    DPS~\cite{chung2022diffusion}               & 15.72 & 0.685 & 0.2597 & 1.222 & 0.8660 & 0.9473 & 0.8972 & 0.9551 & 18.97 & 0.731 & 0.1953 \\
    ILVR~\cite{choi2021ilvr}                    & 16.50 & 0.704 & 0.2360 & 0.866 & 0.9418 & 0.9710 & 0.9538 & 0.9863 & 20.62 & 0.763 & 0.1548 \\
    Ours                                        & 22.80 & 0.854 & 0.0954 & 0.574 & 0.9746 & 0.9835 & 0.9781 & 0.9945 & 25.46 & 0.885 & 0.0566 \\
  \bottomrule
  \end{tabular}
  }
\end{table}

\begin{table}[tb]
  \caption{Quantitative Evaluation on Inpainting Part with Generation metrics (Toys4k).}
  \label{tab:hssd_gen_quan}
  \centering
  \resizebox{0.7\linewidth}{!}{
  \begin{tabular}{@{}l|ccccc|c@{}}
    \toprule
    Method & CLIP$_\uparrow$ & FD$_{dinov2j,\downarrow}$ & KD$_{dinov2,\downarrow}$ \\
    \midrule
    RePaint~\cite{lugmayr2022repaint}           & 28.60 & 25.86 & 2.043 \\
    SDEdit~\cite{meng2021sdedit}                & 26.40 & 22.98 & 1.650 \\
    BlendedDiffusion~\cite{avrahami2022blended} & 28.60 & 25.94 & 2.257 \\
    MultiDiffusion~\cite{bar2023multidiffusion} & 28.53 & 26.97 & 2.101 \\
    DPS~\cite{chung2022diffusion}               & 28.33 & 31.39 & 2.440 \\
    ILVR~\cite{choi2021ilvr}                    & 28.31 & 16.73 & 1.198 \\
    Ours                                        & 28.38 & 6.675 & 0.549 \\
  \bottomrule
  \end{tabular}
  }
\end{table}

\end{document}